# Development, Optimization, and Deployment of Thermal Forward Vision Systems for Advance Vehicular Applications on Edge Devices

Muhammad Ali Farooq *[1], Waseem Shariff [1], Faisal Khan [1], Peter Corcoran[1]
[1]University of Galway

## ABSTRACT

In this research work, we have proposed a thermal tiny-YOLO multi-class object detection (TTYMOD) system as a smart forward sensing system that should remain effective in all weather and harsh environmental conditions using an end-to-end YOLO deep learning framework. It provides enhanced safety and improved awareness features for driver assistance. The system is trained on large-scale thermal public datasets as well as newly gathered novel open-sourced dataset comprising of more than 35,000 distinct thermal frames. For optimal training and convergence of YOLO-v5 tiny network variant on thermal data, we have employed different optimizers which include stochastic decent gradient (SGD), Adam, and its variant AdamW which has an improved implementation of weight decay. The performance of thermally tuned tiny architecture is further evaluated on the public as well as locally gathered test data in diversified and challenging weather and environmental conditions. The efficacy of a thermally tuned nano network is quantified using various qualitative metrics which include mean average precision, frames per second rate, and average inference time. Experimental outcomes show that the network achieved the best mAP of 56.4% with an average inference time/ frame of 4 milliseconds. The study further incorporates optimization of tiny network variant using the TensorFlow Lite quantization tool this is beneficial for the deployment of deep learning architectures on the edge and mobile devices. For this study, we have used a raspberry pi 4 computing board for evaluating the real-time feasibility performance of an optimized version of the thermal object detection network for the automotive sensor suite. The source code, trained and optimized models and complete validation/ testing results are publicly available at https://github.com/MAli-Farooq/Thermal-YOLO-And-Model-Optimization-Using-TensorFlowLite.

## 1. INTRODUCTION

Thermal imaging has shown superior benefits over the past few years for diversified real-world applications. Thermal imaging when integrated with advanced machine learning and computer vision algorithms can be used for building AI-based image/video analysis systems for various applications such as infrared thermography for disease diagnosis [1], non-destructive testing [2], machine health checks [3], surveillance hardware [4], crop and field monitoring [5], thermal object detection for advance driver assistance systems [6], etc. In this work, we have proposed a smart forward sensing system for providing comprehensive roadside object information to the driver. Such systems can be effectively used for improved situational awareness by generating timely alerts and preventing any possible collisions. As compared to visible or conventional CMOS imaging cameras, the performance of thermal imaging modules is not affected in low lighting scenarios and remains fully operational even in zero lighting conditions. Moreover, thermal cameras emerged as a robust solution in harsh weather circumstances as it has the ability to generate clear images and video data even through smoke, fog, dusty and sandy conditions. The further advantages include strong durability, extended operational temperature ranges, and the wider field of view of thermal cameras which is eventually not in the case of RGB CMOS devices which can be effectively used for increasing the capabilities of more convention "Machine Vision" imaging systems

In this research work which is carried out under the Heliaus project [7], we have proposed and developed a tiny object detection framework for robust detection and classification of stationary as well as moving roadside objects from six different classes i.e., person, car, poles, bike, bus, and bicycle. The key challenge is the robust detection of static and dynamic objects in various environments, in both cluttered and uncluttered views, independent of atmospheric conditions or locale using light-weight CNN architecture. To achieve this goal, we propose the use of a novel adaptation and convergence of the end-to-end YOLO-V5 tiny model [8] to establish a fast yet robust inference network in order to accommodate the vehicle's real-time environment for the detection and classification of objects in thermal imagery. In addition, we have extended the previous work on thermally tuned network optimization [9], [10] by using the Tensorflow

LITE quantization tool [11] and deploying the quantized tiny-YOLO architecture on low-power embedded board raspberry pi 4 for real-time automotive sensor suite feasibility tests. In our previous work [9], [10] we used Nvidia Jetson & Xavier developments board for the purpose of deploying an optimized inference network by using the TensorRT inference accelerator however these 'power-hungry' Edge-GPU systems are not very suitable for next-generation electric vehicles and there is a need for running optimized neural inference networks in real-time on lower-power platforms.

For performing optimal generalization and achieving satisfactory validation results we have trained the tiny network variant on the large-scale public as well as our locally acquired thermal object detection datasets also known as C3I Thermal automotive dataset [12] available at (https://ieee-dataport.org/documents/c3i-thermal-automotive-dataset). This dataset is acquired a using prototype uncooled Long Wave Infrared (LWIR) ice-cube thermal imaging module. The overall data is recorded in 640x480 resolution with 30 FPS configuration. Fig.1 depicts the sample thermal frames acquired in different weather and atmospheric conditions along with LWIR thermal imaging module.

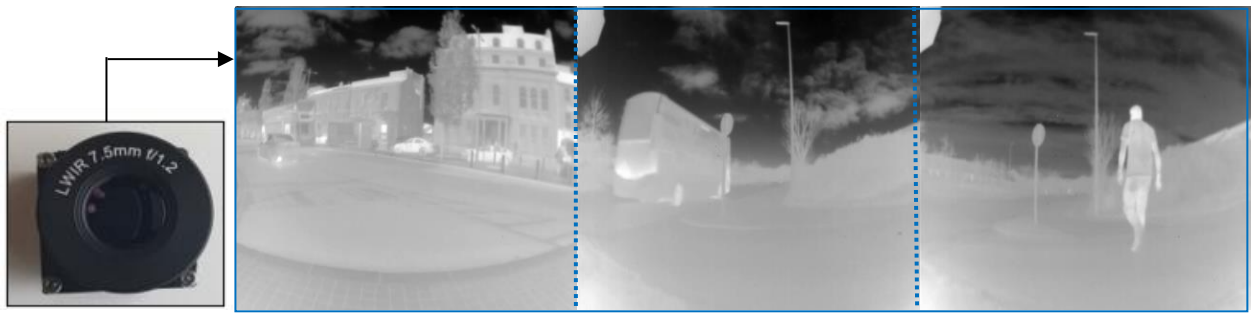

Figure 1. Prototype 640x480 LWIR thermal camera on the left side, and right-hand side show three thermal frames captured in the evening time, nighttime, and daytime.

## 1.1 Research Contributions

The major contributions of this work are as follows.

1. We examine the effectiveness of applying fine-tuning approach by using transfer learning of YOLO-v5 nano network weights originally trained on an RGB COCO dataset on thermal images acquired from the four different public as well as locally acquired novel C3I [12] datasets.

2. Performing network optimization/quantization using state-of-the-art (SoA) neural inference accelerator for deploying the trained network structure on a low-power single board edge device.

The rest of the paper is structured as follows, section 2 presents the background by discussing previously published studies, section 3 discusses the proposed algorithm along with the training/ tuning methodology, section 4 shows the qualitative and quantitative experimental results, and further discusses the thermally tuned network optimization and deployment results on raspberry pi 4 computing board, and section 5 present the conclusion and future possibilities in this domain.

## 2. BACKGROUND/ LITERATURE REVIEW

In recent years we have witnessed rapid progress in terms of the precise detection rate and accuracy of object detection algorithms based on deep learning methods such as convolution neural networks (CNN). End-to-End CNN has significantly decreased the missed detection rate of multispectral objects in the thermal spectrum [13]. Two-stage object detection networks such as Faster RCNN [14], and MSDS RCNN [15] have shown superior performance on diversified real-world applications [16],[17], [18] however these networks require higher computational time [19]. To cater this problem, single-stage detectors such as YOLO [20], and SSD [21] were proposed to achieve a faster inference rate. Huang et al. [22] have tested various patterns of object detection frameworks by modifying different components and parameters to find the best possible configurations for specific scenarios, e.g., deployment on mobile devices. El Ahmar, Wassim A., et al [23] proposed a multi-object detection and tracking method on thermal data along with their corresponding visible data. Their experimental result shows that transfer learning is more effective even on different data modalities such as thermal data when used to train a detector on a dataset from the same spectrum as the initial weights. In this work, we have utilized single stage detector Yolo-v5 [8] for designing a tiny-YOLO multi-object detection system using thermal imaginary for ADAS which is derived from our previous published work [9], [10]. Yolo-V5 network variants employ a single neural network to process the entire picture, then separate it into components and predict bounding boxes and probabilities for each component. These bounding boxes are weighted by the estimated probability. It makes predictions after only one forward propagation runs through the neural network. Finally, it then delivers outputs after non-max suppression which ensures that the object detection algorithm only identifies every single object once.

SoA CNN-based object detectors are computationally expensive and memory hungry thus it is a challenge to deploy these models in edge devices, without performing model optimization [24]. CNN model optimization allows us to perform on-chip machine learning with limited computational power. The core advantages of CNN model optimization include low latency, no requirement of internet connectivity, and lastly, the lightweight model structure is beneficial for an efficient inference network and low power consumption. We can find various open-source optimization tools such as TensorFlow Lite [11], Tensor RT [25], OpenVINO [26], etc. which are most commonly used for performing network quantization thus allowing us to rapidly deploy quantized CNN models on edge devices, mobile platforms such as Android and IOS and cloud structures.

## 3. PROPOSED ALGORITHM AND OPTIMIZATION METHODOLOGY

This section will discuss the proposed working methodology and will further present the optimization methodology adapted to optimally tune the tiny network variant of Yolo-V5 [8] on thermal imaging data that consists mainly of convolutional layers with the least number of model parameters and originally trained on the COCO dataset. Fig.2 shows the block diagram presentation of the proposed algorithm.

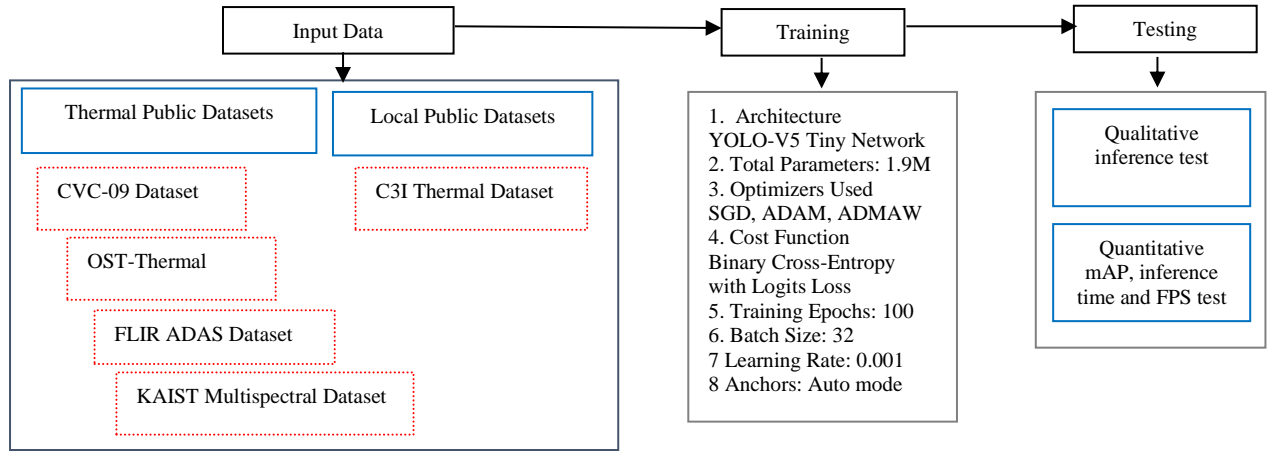

Figure 2. Block diagram of proposed YOLO-tiny thermal object detection system.

As it can be observed from Fig.2 we have used 5 different datasets for the training purpose among which 4 are public datasets [27], [28], [29], [30] whereas the fifth dataset is our recorded thermal C3I thermal dataset [12]. The main reason for including a variety of public datasets is to produce enough diversity and variation in the training set. Also during the training phase, we have used various data augmentation techniques which involve data cropping, scaling, mosaic augmentation, random flipping, and rotation for optimal convergence and avoiding model over-fitting on thermal data. During the training phase, three different SoA optimizers (SGD, ADAM, and ADAMW) were used to achieve the best training results. It is important to mention that rather than training the network from scratch we have used the transfer learning approach to save time and computational cost. The overall training data is divided in the ratio of 72%-28% with a total of 11,160 images as the training set and 3,213 images as the validation set. The recent versions of the YOLO framework require anchors box tuning for the custom training datasets. Anchor boxes are defined as a group of predetermined bounding boxes with a specific height and width. These boxes are often selected based on object sizes in custom training datasets and are defined to capture the scale and aspect ratio of various object classes to be detected. Since YOLO predicts bounding boxes implicitly but as displacements from anchor boxes, the selection of appropriate anchors is a crucial step. YOLO-V5 comes with the pre-built auto-anchor algorithm. The auto-anchor algorithm works in a way that the provided anchors are checked to see how well they fit the data before the training process, and if they don't, the algorithm recalculates them. The model is then trained using new, best anchors. It appears to be an extremely beneficial function, especially on smaller custom datasets. Fig.3 shows the best anchors results in our case.

```
AutoAnchor: 4.81 anchors/target, 0.999 Best Possible Recall (BPR). Current anchors are a good fit to dataset ✅
Image sizes 640 train, 640 val
Using 8 dataloader workers
Logging results to runs/train/exp
Starting training for 100 epochs...
```

Figure 3. Best computed possible recall rate (BPR) and anchors/target using auto-anchor algorithms.

### 3.1 CNN Model Optimization

The main goal of CNN architecture optimization is to minimize the number of model parameters. The reduced parameter version of the DNN model translates into fewer computations and faster inference times. This can be achieved by utilizing various methods which include parameter removal, parameter search, parameter decomposition, and parameter quantization. In this study, we have used the TensorFlow Lite model quantization tool [11] for performing tiny-YOLOv5 variant optimization. TensorFlow Lite is an open-source, product-ready, cross-platform deep learning framework that converts a custom-trained model to a special format that can be optimized for speed or storage. The quantized version of the model generated through TF Lite results in being lightweight and requires less computational power which is advantageous to rapidly deploying the DNN architecture to edge-embedded boards and mobile devices. In this work, we have selected the raspberry pi 4 model B for the optimized thermal-tiny version deployment. The credit card size raspberry pi 4 comes with a Broadcom BCM2711 chip having Quad core Cortex-A72 (ARM v8) 64-bit SoC with a clock rate of 1.5GHz, and 4GB of ram [31]. Fig.4 shows a raspberry pi 4 board placed in a silver metal case and equipped with a 5-volt cooling fan powered from general purpose input-output (GPIO) pins to prevent the microcontroller unit from heating up while running the inference tests.

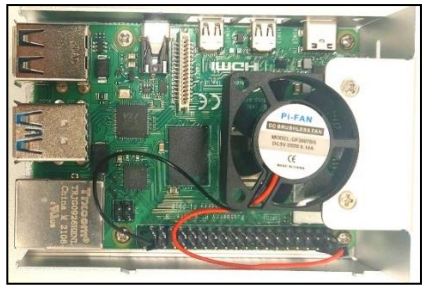

Figure 4. Raspberry pi 4 model B board mounted in the silver metal case and has 5-volt DC cooling fan taking power from GPIO pins.

## 4. EXPERIMENTAL RESULTS

This section will review the training and validation results of the tiny-YOLO object detection algorithm on thermal data. The complete training and testing phase was carried out on work station machine having Intel(R) Core(TM) i7-9700K CPU, 64 GB DDR4 RAM, and a dedicated Nvidia R30360 GPU card. Moreover, we have used the PyTorch and TensorFlow deep learning tools for carrying out the complete experimental work.

### 4.1 Training Results

As mentioned earlier the tiny variant has the smallest number of model parameters = 1.9 million as compared to other network variants of Yolo-v5 which include small, medium, large, and x-large variants. In our previous published work, we have already performed detailed experimental and comparison analysis of Yolo-v5 bigger models [8] on thermal data therefore in this work we have primarily focused on the lightweight yet fastest network variant. The architecture was initially trained and tested on the COCO dataset with an image size of 640 where it achieved the best mAP score of 28% and floating point operation (FLOP) rate of 4.5 billion/second. The network is trained using the same set of hyperparameters as shown in Fig.2 however three different optimizers SGD, ADAM, and ADAMW are used during each training phase. It is important to mention that rather than training the network from scratch by unfreezing all the network layers we have used transfer learning to optimally converge the pre-trained nano model weights on thermal data. Fig.5 shows the comparison of training graphs (mAP, precision, and recall) and loss graphs while using three different optimizers. As it can be seen from Fig.5, the SGD optimizer has performed considerably better as compared to ADAM and ADAMW in terms of various accuracy and loss metrics. Thus, SGD optimized trained model was shortlisted for testing/ validation purposes. Fig.6 shows the training mAP and F1-score results on six different classes which are extracted using the SGD optimizer.

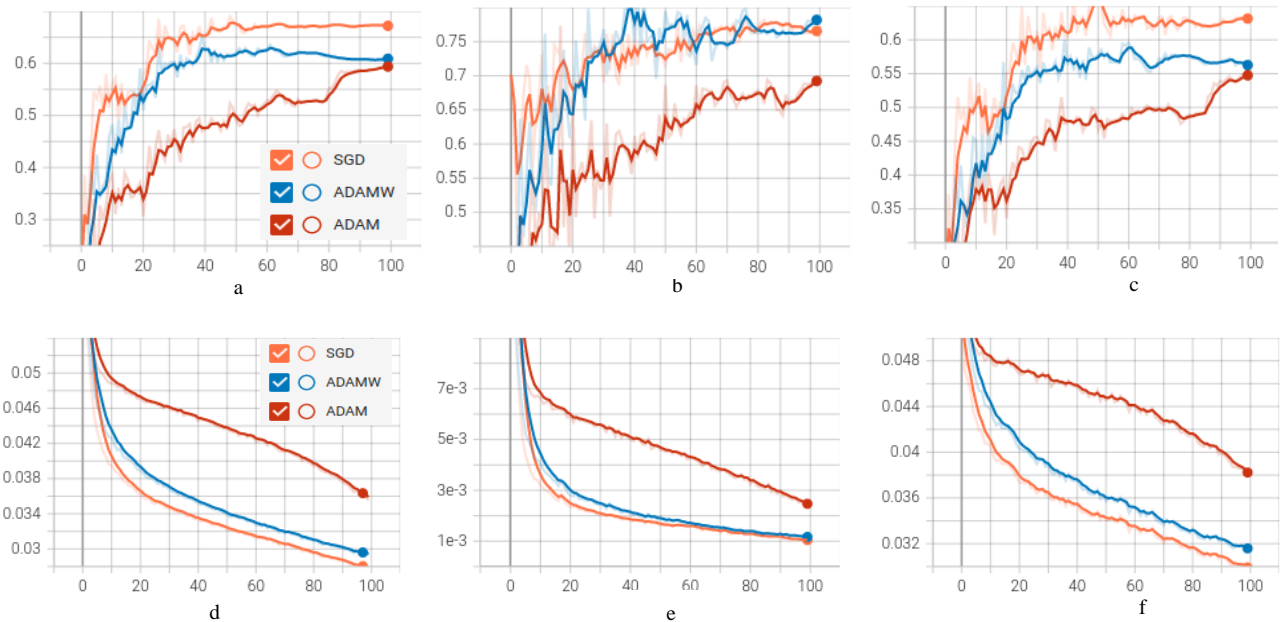

Figure 5. Comparison of training and loss results using SGD, ADAMW, and ADAM optimizer, a) mAP (highest score = SGD = 68.9%), b) precision ( highest score = ADAMW = 78.1%), c) recall (highest score = SGD = 64.2%), d) box-loss (lowest value = SGD = 0.0285), e classification-loss (lowset value SGD and ADAMW = 0.001), f) object loss (lowset value = SGD = 0.0282). .

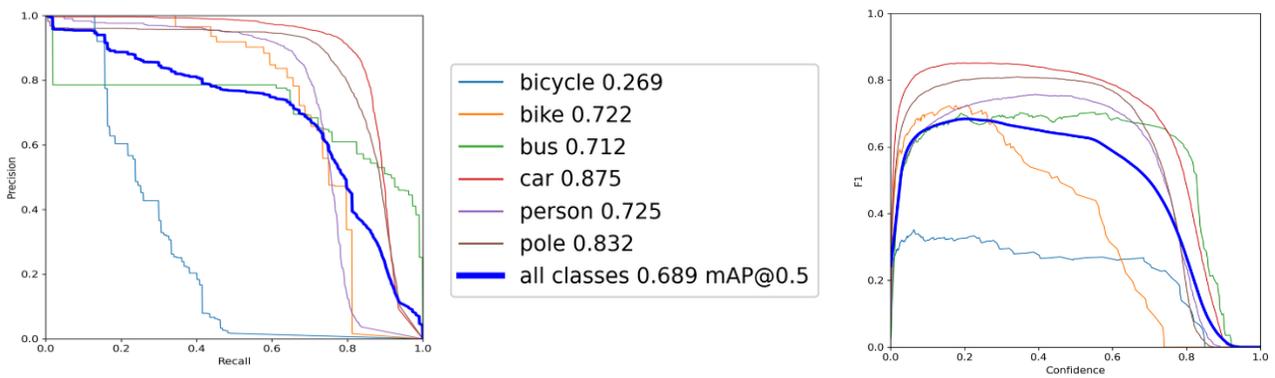

Figure 6. Individual classwise training results of the Yolo-v5 nano model using SGD optimizer, the right side shows the mAP score of six different classes with an overall mAP of 68.9%, and left side shows the F1 curve of six different classes with the confidence value that optimizes the precision and recall is 0.201, corresponding to the maximum F1 value of 68%.

### 4.2 Validation/Testing Results

After the training phase, we tested the performance of the thermal tiny-YOLO multi class object detection (TTYMOD) model on large-scale unseen test thermal data. The overall inference test was carried out on more than 30,000 thermal frames collected from both public as well as locally acquired datasets. Table 1 shows the quantitative validation results on a test image set of 500 thermal frames with complex scenes such as overlapping classes, multiple objects of different classes in a single frame, and various environmental conditions. Also, we have used various combinations of Confidence thresholds and Intersection of Union (IoU) threshold values to check the efficiency of thermally tuned nano architecture with a batch size of 64 and image size of 640x480 pixels.

Table 1. Quantitative validation results on thermal test data (best values highlighted in green color).

| Experiment No | Confidence and Iou Threshold | mAP (all classes) % | Classwise mAP % | Average inference time/frame (ms) | Minimum inference time (ms) | FPS |
|---|---|---|---|---|---|---|
| 1 | 0.2, 0.2 | 55.4 | Bicycle: 35.8<br>Bike: 40.8<br>Bus: 41.5<br>Car: 81.3<br>Person: 73.1<br>Pole: 60.1 | 4 | 2 | 250 |
| 2 | 0.4, 0.4 | 56.4 | Bicycle: 36.6<br>Bike: 47.8<br>Bus: 41.4<br>Car: 80.1<br>Person: 71.9<br>Pole: 60.8 | 5 | 2 | 200 |
| 3 | 0.3, 0.6 | 53.9 | Bicycle: 29.6<br>Bike: 40.3<br>Bus: 39<br>Car: 81.0<br>Person: 72.9<br>Pole: 60.7 | 4 | 2 | 230 |

From Table 1 we can observe that by selecting the confidence threshold and IoU threshold value of 0.4 we have achieved the best mAP scores however in terms of inference speed and FPS scores the selected combination of confidence and IoU thresholds of 0.2 has achieved the best FPS results. Fig.7 shows the qualitative inference results on various complex thermal frames by using the confidence threshold and IoU threshold value of 0.4.

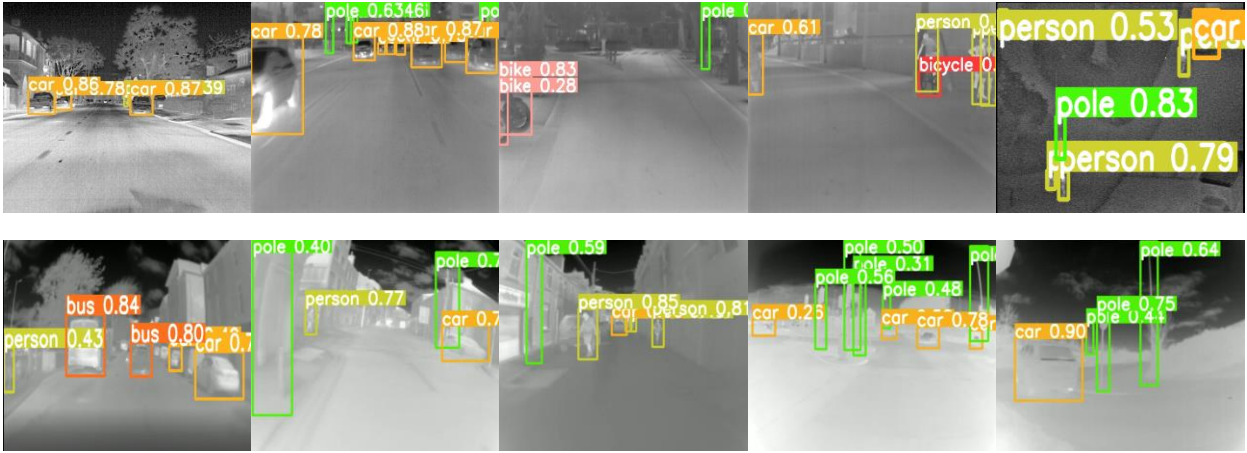

Figure 7. Robust inference results on 10 different complex frames (the above row shows results on public datasets whereas the below row shows results on the locally acquired dataset) with multiple objects of the same and different classes, overlapping/ cluttered classes, and different atmospheric and environmental conditions.

The complete inference results along with the testing results are available on the GitHub page and can be downloaded from the provided link. (https://github.com/MAli-Farooq/Thermal-YOLO-And-Model-Optimization-Using-TensorFlowLite).

### 4.3 Network Optimization Results

The last phase of the experimental work includes model optimization using the TensorFlow Lite tool. Fig.8 shows the complete model quantization pipeline. As it can be observed from Fig.8 that pytorch trained model weight is first converted to TensorFlow weight format which is then converted to TF Lite format using TF Lite converter python API in FlatBuffer format. The optimized FlatBuffer weights are then imported to the raspberry pi 4 computing board through a cloud sharing network. To successfully run the optimized model weights on the edge computing board, all the necessary python libraries are installed as per TFLite documentation. In the next stage, we have run the inference test on thermal test data.

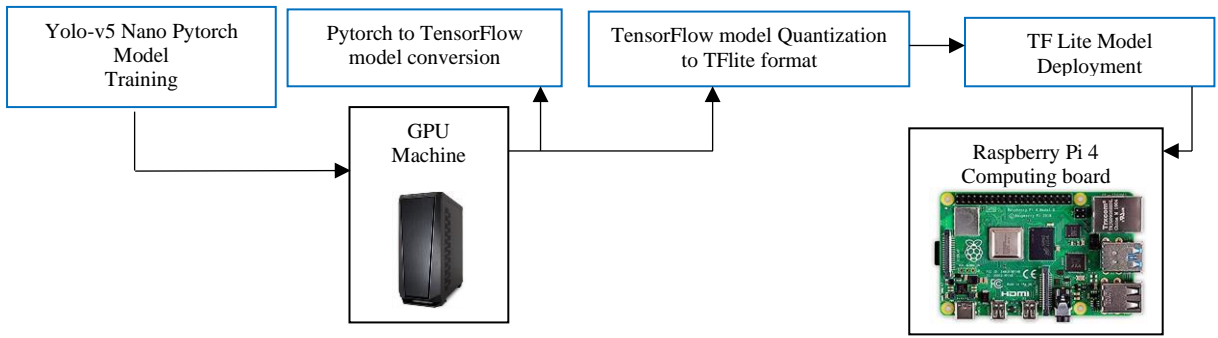

Figure 8. TTYMOD network optimization pipeline for deployment on the single board edge device.

Fig.9 shows the inference results on numerous complex frames with multiple objects whereas Table 2 shows the quantitative validation and inference speed results on the raspberry pi 4 computing board.

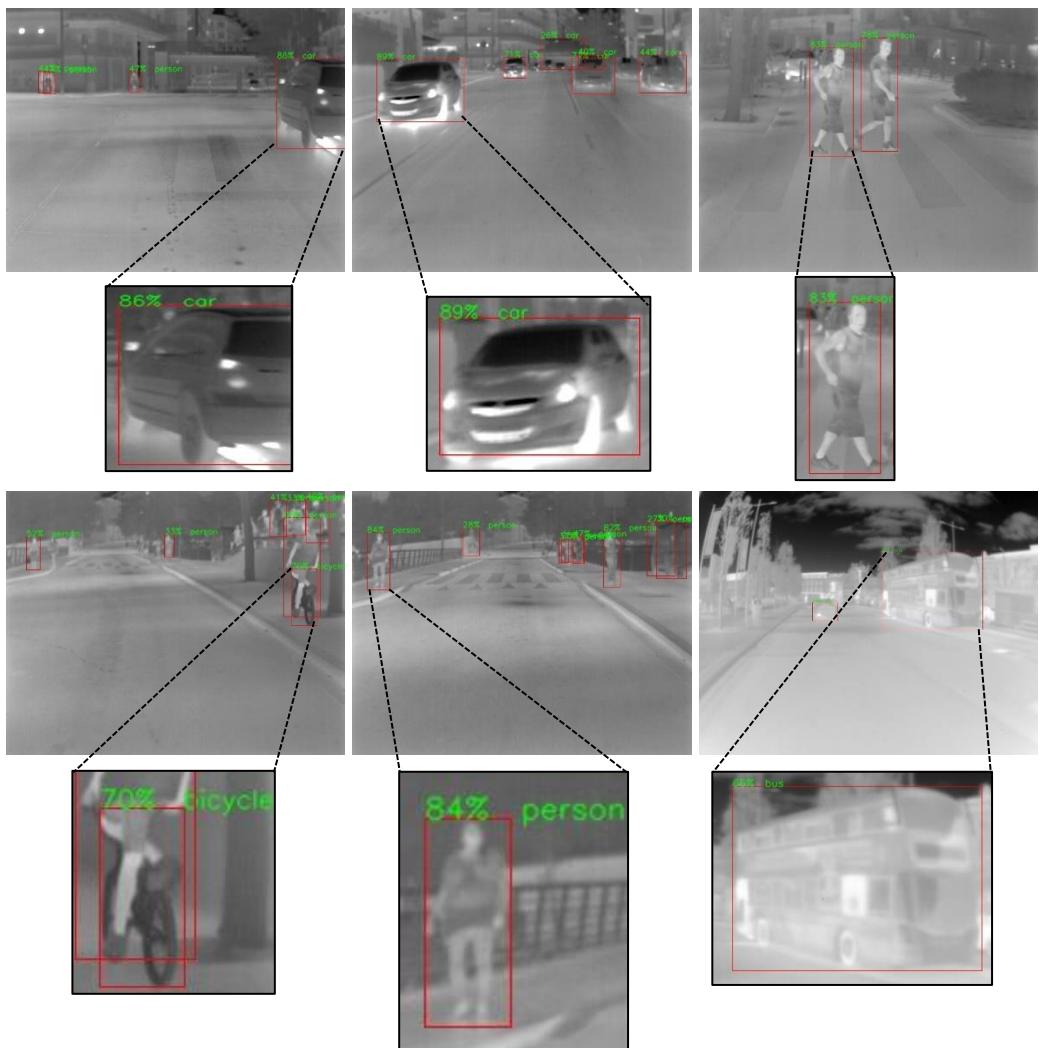

Figure 9. Inference results on six different thermal frames showing different class objects using an optimized version of TTYMOD deployed on raspberry pi 4.

Table 2. Quantitative validation results using the optimized network version

| Image size | Confidence and Iou Threshold | mAP (all classes) % | Classwise mAP % | Average inference time/frame (sec) | Minimum inference time (sec) | FPS |
|---|---|---|---|---|---|---|
| 416 x 416 | 0.2, 0.2 | 55.4 | Bicycle: 35.8<br>Bike: 40.8<br>Bus: 41.5<br>Car: 81.3<br>Person: 73.1<br>Pole: 60.1 | 0.8 | 0.6 | 1 |
| 128 x 128 | 0.2, 0.2 | 55.4 | Same as 416X416 image results | 0.4 | 0.3 | 2.5 |

## 5. CONCLUSION AND FUTURE WORK

In this study, an algorithm for a robust thermal multi-class object detection is proposed based on transfer learning and fine-tuning a single-stage tiny-YOLO network on thermal data which can be efficiently used as an optics-based forward sensing system for out-cabin vehicular technology. The experimental outcomes show that the SGD optimizer along with the selection of appropriate network hyper-parameters helps in achieving optimal training results with an overall training mAP of 68.9%. The efficacy of the optimally trained CNN network is validated on diversified test data using both qualitative and quantitative metrics. The TTYMOD achieves the maximum validation mAP of 56.4% with 200 FPS on the GPU computing system. Further, the TTYMOD is deployed on ARM processor based raspberry pi 4 computing board, by performing network optimization using the SoA TensorFlow Lite tool. The minimum inference time required on raspberry pi 4 is 0.3 seconds per frame using the frame size of 128.

At this stage we have yet not achieved very good FPS on the raspberry-pi 4 board however as the possible future directions we can focus on other inference optimization techniques such as model parameter decomposition, and post-training static quantization. Moreover, these models can be deployed on more advanced single board edge computing devices such as NVIDIA Jetson AGX Xavier and Nvidia Jetson TX2 board in low power mode to achieve more practical FPS for real-world applications.

## ACKNOWLEDGMENT

This research work is carried out under the EU Heliaus project. This project has received funding from the ECSEL Joint Undertaking (JU) under grant agreement No 826131. The JU receives support from the European Union's Horizon 2020 research and innovation program and National funding from France, Germany, Ireland (Enterprise Ireland International Research Fund), and Italy". The authors would like to acknowledge the contributors of all the public datasets for providing the image resources to carry out this research work and ultralytics for sharing the YOLO-V5 Pytorch version.

## AUTHORS' BACKGROUND

| Your Name | Title* | Research Field | Personal website |
| --- | --- | --- | --- |
| Dr Muhammad Ali Farooq | Post-Doctoral researcher | Computer Vision, Machine Learning, Edge Computing, Machine vision, Camera calibration | LinkedIn: https://www.linkedin.com/in/muhammad-ali-farooq-phd-876235a1/ |
| Waseem Shariff | Ph.D. candidate | Computer Vision and Machine Learning | LinkedIn: https://www.linkedin.com/in/waseem-shariff-997534141/ |
| Faisal Khan | Optical Engineer & Ph.D. candidate | Computer Vision and Machine Learning | LinkedIn: https://www.linkedin.com/in/faisal-khan-1b4a78123/?originalSubdomain=ie |
| Dr Peter Corcoran | Full professor | Computer imaging, machine learning, digital image processing, camera calibration | LinkedIn: https://www.linkedin.com/in/pcor00/ |